\documentclass[10pt]{article}
\usepackage{spconf,amsmath,epsfig,amsfonts,multirow,marvosym}
\usepackage{color}
\graphicspath{{./images/}}
\let\OLDthebibliography\thebibliography
\renewcommand\thebibliography[1]{
  \OLDthebibliography{#1}
  \setlength{\parskip}{0pt}
  \setlength{\itemsep}{0pt plus 0.3ex}
}

\begin{document}\sloppy

\def\x{{\mathbf x}}
\def\L{{\cal L}}

\title{Scene Graph based Fusion Network for Image-Text Retrieval}
%
\name{Guoliang Wang\textsuperscript{1}, Yanlei Shang\textsuperscript{1} and  Yong Chen\textsuperscript{1}\Letter}
\address{\textsuperscript{1}Beijing University of Posts and Telecommunications, China\\
\{wgl2019,shangyl\}@bupt.edu.cn,~~~~ alphawolf.chen@gmail.com}
\maketitle

\begin{abstract}
	A critical challenge to image-text retrieval is how to learn accurate correspondences between images and texts. 
	Most existing methods mainly focus on coarse-grained correspondences based on co-occurrences of semantic objects, while failing to distinguish the fine-grained local correspondences. In this paper, we propose a novel \textit{Scene Graph based Fusion Network} (dubbed SGFN), which enhances the images'/texts' features through intra- and cross-modal fusion for image-text retrieval. 
	To be specific, we design an intra-modal hierarchical attention fusion to incorporate semantic contexts, such as objects, attributes, and relationships, into images'/texts' feature vectors via scene graphs, 
	and a cross-modal attention fusion to combine the contextual semantics and local fusion via contextual vectors. 
	Extensive experiments on public datasets Flickr30K and MSCOCO show that our SGFN performs better than quite a few SOTA image-text retrieval methods.
\end{abstract}
\begin{keywords}
	Scene Graph, Hierarchical Attention, Contextual Vectors, Image-Text Retrieval
\end{keywords}

\section{Introduction}
\label{sec:intro}

Image-text retrieval is one of the fundamental tasks in the field of vision and language~\cite{Li2022VisionLanguageIT}. 
Its goal is to effectively retrieve the most similar samples to its content from the database of image (text) modality given a query of text (image) modality. 
The biggest challenge is to narrow the semantic gap between cross-modal data for accurate similarity of image-text pairs.

To tackle such issue, many solutions have been proposed. 
Early methods are usually to map the global feature vectors of images and texts into a common latent space, in which the more similar the semantics between images and texts are, the closer they are located~\cite{wang2018learning,radford2021learning}. 
Nevertheless, when the image or text modality involves complex scenes (e.g., containing multiple entities), the retrieval performances of these methods are usually unsatisfactory. 

Thereafter, more and more studies are paying attentions to cross-modal local correspondences by extracting local features. 
To name a few,  
Lee \textit{et al.}~\cite{lee2018stacked} employed a cross-modal attention mechanism to combine all image regions with different weights to represent each textual word, and vice versa. 
Following this work, Wang \textit{et al.}~\cite{ijcai2019p526} further used positional embeddings for regions to guide the correspondence learning. 
Liu \textit{et al.}~\cite{liu2020graph} adopted polar coordinates to model the positional relationship of regions in an image. 
Besides, there are also some other approaches that consider using intra- and cross-modal fusions to bridge the feature gap between different modalities. For example, Wei \textit{et al.}~\cite{wei2020multi} built a network that exploits an attention mechanism to achieve intra-modal fusion and cross-modal fusion. 
Ji \textit{et al.}~\cite{ijcai2021p106} proposed a step-by-step hierarchical alignment fusion network to infer the similarity of image-text pairs.

\begin{figure}[tbp]
\centering
\includegraphics[width=1.0\columnwidth]{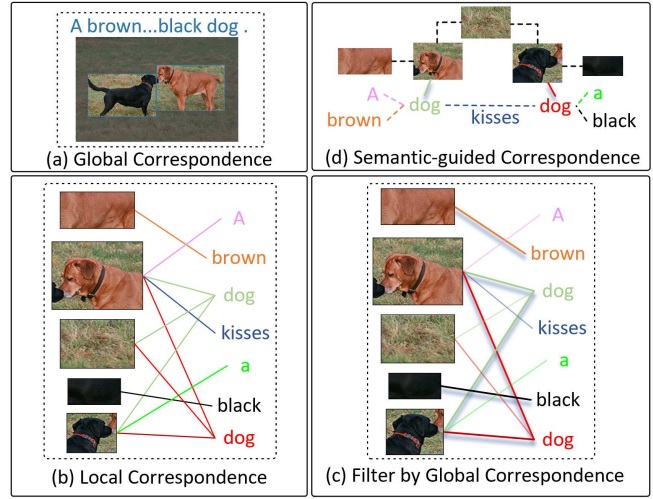}
\caption{Illustrations of global/local
	correspondences between image regions and textual words: (a) Global correspondence focuses on shared overall pictures. 
	(b) Local correspondence focuses on alignments between local semantic components. 
	(c) Global correspondences enhance the important matches in local correspondences, and the corresponding lines are thickened. 
	(d) The two words ``\textit{dog}" and two regions ``\textit{dog}" can be easily distinguished after combining specific contexts (e.g., attributes, relationships) in the scene graph.}
\label{figure1}
\end{figure}

Although these methods have reported encouraging cross-modal retrieval performance, there still exist some problems: 
(1) they neglect the potential relationships between global and local correspondences. As can be seen from Fig.~\ref{figure1}(a) \& (b) to Fig.~\ref{figure1}(c), the global correspondences can effectively enhance the important local correspondences. 
(2) when performing local correspondence and fusion, they did not consider the ambiguity of these correspondences, as shown in Fig.~\ref{figure1} (b) \& (c), there are many-to-many correspondences between two words ``\textit{dog}'' and two regions ``\textit{dog}''. Integrating semantic contexts (e.g., attributes, relationships, relative positions, etc.) of the modality itself into the local correspondences can effectively distinguish them, as illustrated in Fig.~\ref{figure1}(d).

To learn the fine-grained semantics between images and texts, we propose a novel \textit{\underline{S}cene \underline{G}raph based \underline{F}usion \underline{N}etwork} (SGFN) that contains two critical intral- and inter-modal fusion components for more accurate representations of images and texts. 
Concretely, SGFN explicitly builds objects, attributes, and relationships via scene graphs, and exploits hierarchical attention to selectively incorporate them into images' and texts' feature vectors. 
Besides, SGFN exploits empty global agent vectors for image and text to dynamically acquire the overall features during intra-modal fusion. Note that the global agent vectors can be used to form contextual vectors, which contain shared object's features and filter important local correspondences for cross-modal fusion. 
To the end, the updated local feature vectors are used to compute the similarities of image-text pairs with the global and local alignments.

To sum up, the major contributions are listed as follows:
\begin{enumerate}
	\item Propose a scene graph based fusion network (SGFN), which constructs scene graphs for images and texts and then performs hierarchical attention fusions to selectively incorporate semantic contexts, e.g., objects, attributes or relationships, into their intra-modal features.
	
	\item Form a contextual vector through a global agent vector and use it to guide cross-modal fusion. To the best of our knowledge, this is the first trial that global agent vectors dynamically guides cross-modal fusions.
	
	\item Conduct image-text retrieval experiments on public datasets Flickr30K and MSCOCO to show SGFN's advantages over SOTA methods and verify the effectiveness of the intra- and inter-modal fusion components for image-text retrieval via ablation studies.
\end{enumerate}

\section{Related Work}

\textbf{Scene Graph.} Scene graph can be used to describe objects and their attributes and relationships in images or texts, which is gradually adopted in image-text retrieval tasks. 
For example, SGM~\cite{wang2020cross} transforms the image-text retrieval task into a matching task of two scene graphs but ignore the visual feature itself. 
LGSGM~\cite{Nguyen2021ADL} incorporates global and local correspondences into the scene graph matching task but fails to distinguish the importance of different kinds of nodes in scene graph. 
Unlike them, our SGFN leverages scene graph to model objects, attributes, and their relationships, and designs hierarchical attention fusion to integrate them into their intra-modal feature vectors.

\noindent\textbf{Attention Mechanisms.} Attention mechanism can be used to simulate the behavior of human beings to selectively focus on important parts of the modality, and has been widely used in image-text retrieval task~\cite{nagrani2021attention,zhou2021deepvit,chen2021crossvit}. In~\cite{lee2018stacked}, a stacked cross attention mechanism is adopted to discover the latent alignments  between regions and words but ignores the intra-modal relationship. 
In~\cite{wu2019learning}, they implement the self-attention mechanism by Transformer model but ignores the inter-modal relationship. MMCA~\cite{wei2020multi} proposes to use a unified transformer to achieve cross-modal fusion and intra-modal fusion while ignoring the relationship between global and local correspondences. Different from existing methods, the cross-attention proposed in this paper uses contextual vectors to filter local features for cross-modal fusions.

\section{scene graph based fusion network}

\begin{figure*}[ht]
\centering
\includegraphics[width=0.95\linewidth]{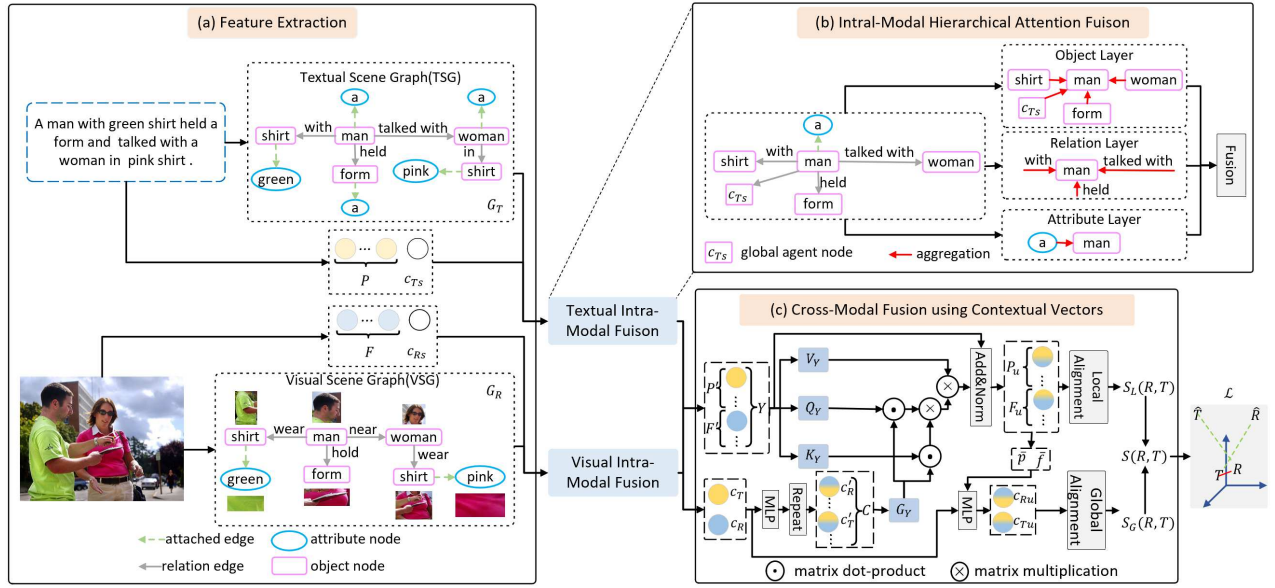}
\caption{The framework of our proposed Scene Graph based Fusion Network (SGFN).}
\label{figure2}
\end{figure*}

The overall architecture of our SGFN is drawn in Fig.~\ref{figure2}, elaborated step-by-step as below.

\subsection{Feature Extraction for Images/Texts}

\textbf{Visual Representation.} We extract $n$ region-level visual features for each image using FasterRCNN pre-trained on Visual Genomes\footnote{https://visualgenome.org/}, and add a fully connected layer to transform their dimensions into $d$ dimensions, i.e., $F=\{f_1,f_2,..,f_n\}\in \mathbb{R}^{n\times d}$. Then, we leverage the RelTR~\cite{cong2022reltr} tool and word embeddings to generate the corresponding scene graph $G_R=\{O_R,E_R,A_R,M_R\}$, where the object set is $O_R=\{o_{R1},..,o_{Rn}\}\in \mathbb{R}^{n\times d}$, the relationship set is $E_R=\{e_{R1},...,e_{Ri1}\}\in \mathbb{R}^{i_1\times d}$, the property set is $A_R=\{a_{R1},...,a_{Rj1}\}\in \mathbb{R}^{j_1\times d}$ and the adjacency matrix is $M_R \in \mathbb{R}^{n\times n}$. Define the global agent vector as $c_{Rs}=[1,...,1]\in \mathbb{R}^{1\times d}$.

\noindent\textbf{Textual Representation.} Given a sentence, we input the one-hot vectors of $m$ words into the pre-trained bi-GRU to get the representation of each word, i.e., $P=\{p_1,...,p_m\}\in \mathbb{R}^{m\times d}$, which are used to represent nodes' embeddings in the scene graph. Then, we use StanfordCoreNLP to generate the scene graph $G_T=\{O_T,E_T,A_T,M_T\}$, where the object set is $O_T=\{o_{T1},...,o_{i2}\in \mathbb{R}^{i_2\times d} \}$, the relationship set is $E_T=\{e_{T1},...,e_{Tj2}\}\in \mathbb{R}^{j_2\times d}$, the property set is $A_T=\{a_{T1},...,a_{Tk2}\}\in \mathbb{R}^{k_2\times d}$ and the adjacency matrix is $M_T\in \mathbb{R}^{m\times m}$. The global agent vector that defines the text modality is $c_{Ts}=[1,...,1]\in \mathbb{R}^{1\times d}$.

\subsection{Intra-Modal Hierarchical Attention Fusion}

The scene graph is a heterogeneous graph with object nodes, attributes nodes, and their linked edges, as shown in Fig.~\ref{figure2}(a). Considering the heterogeneity, we propose a layered attention fusion consisting of Object layer, Attribute layer and Relation layer, as shown in Fig.~\ref{figure2}(b). The purpose is to study the influence of different types of nodes so that the fusion is more targeted. For the global agent vector, we model a global agent node, which is connected to all nodes.

\noindent\textbf{Textual Intra-Modal Fusion.} We define the initial values of the nodes $h_i=p_i, i\in [1,m], H=\{h_1,...,h_m,c_{Ts}\}$. The object node $i$ in the scene graph is taken as an example to illustrate its update strategy on the three layers. The feature vector of node $i$ is $h_i$, the adjacent object node set is $N_i$, the adjacent edge set is $E_i$, and the adjacent attribute set is $A_i$.

The update rule for node $h_{i}$ in the Object Layer is formulated as $h'_{iobj}$: 
\begin{eqnarray}
    \alpha_{ij}=\frac{exp(LeakyRelu([Wh_i||Wh_j])}{\sum_{k\in N_i}exp(LeakyRelu([Wh_i||Wh_k]))}, \\
    h'_{iobj}=\delta(\frac{1}{K}\sum_{K=1}^K\sum_{j\in N_i}\alpha^k_{ij}W^kh_j),
\end{eqnarray}
where $W$ is a learnable matrix.

The update rules for node $h_{i}$ in the Relation Layer and the Attribute layer are defined as $h'_{irel}$ and $h'_{iatt}$:
\begin{eqnarray}
    h'_{irel}=MLP(\frac{\sum_{j\in E_i}h_j}{|E_i|}),\\
    h'_{iatt}=MLP(\frac{\sum_{j\in A_i}h_j}{|A_i|}). 
\end{eqnarray}

 Three feature vectors represent the information that the object node $i$ needs to pay attention to at three layers, which are fused according to:
\begin{eqnarray}
    h'_i=\frac{e^{\beta}}{e^{\beta}+e^{\alpha}}(h'_{irel}+h'_{iatt})+\frac{e^{\alpha}}{e^{\beta}+e^{\alpha}}h'_{iobj}\label{fusion},
\end{eqnarray}
where $\alpha$ and $\beta$ are trainable parameters.
Finally, we can get $P'=\{p'_1,...,p'_m\}$ with $p'_{i}=h'_{i}$ ($i=1,\cdots,m$) and the global agent node $c_T$ registering modality information.

\noindent\textbf{Visual Intra-Modal Fusion.} Similarly, we get the feature vectors $F'=\{f'_1,...,f'_n\}$ and the global agent node $c_R$.

\subsection{Cross-Modal Fusion using Contextual Vectors}
\noindent\textbf{Cross-Model Fusion.} We adopt a MLP to learn the information interaction between the global agent vectors $c_R$ and $c_T$, thereby forming contextual vectors $c_R'$ and $c_T'$:
\begin{equation}
    \left[c_R^\prime\parallel c_T^\prime\right]=MLP\left(\left[c_R\parallel c_T\right]\right),
\end{equation}
where the contextual vectors pay more attention to the information shared by two modalities to guide cross-modal fusion.

As shown in Fig.~\ref{figure2}(c), we stack the regional features and the word embeddings to obtain $Y=\left[F';P'\right] \in \mathbb{R}^{(n+m)\times d}$. In order to make $c'_R$ and $c'_T$ scale vectors of $F'$ and $P'$, repeat $c'_R$ and $c'_T$ $m$ and $n$  times to get $C'_T\in \mathbb{R}^{m\times d}$ and $C'_R\in \mathbb{R}^{n\times d}$. Similarly, we get $C=\left[C'_R; C'_T\right]$. $K_Y$, $Q_Y$, $V_Y$ and $G_Y$ are formulated as follows:
\begin{eqnarray}
    K_Y=YW^K=\left[F'W^K; P'W^K\right]=\left[K_R; K_T\right],\\
    Q_Y=YW^Q=\left[F'W^Q; P'W^Q\right]=\left[Q_R; Q_T\right],\\
    V_Y=YW^V=\left[F'W^V; P'W^V\right]=\left[V_R; V_T\right],\\
    G_Y=CW^G=\left[C'_RW^G; C'_TW^G\right]=\left[G_R; G_T\right],
\end{eqnarray}
where $W^K$, $W^Q$, $W^V$, and $W^G$ are learnable parameters.

Then, $Y_u$ is formulated as:
\begin{equation}
    \begin{split}
        Y_u=\left[F_u;P_u\right]=softmax(\frac{(G_Y Q_Y)(G_Y K_Y)^T}{\sqrt{d}})V_Y+Y.
    \end{split}
\end{equation}
We re-divide $Y_{u}$ into $F_{u}$ and $P_{u}$, then pass it through an average pooling layer to get the overall representation $\overline{f}\in \mathbb{R}^{1\times d}$ and $\overline{p}\in \mathbb{R}^{1\times d}$. They are used to update contextual vectors with the following equations:
\begin{eqnarray}
    c_{Ru}=MLP(c_R||\overline{f}),c_{Tu}=MLP(c_T||\overline{p}).
\end{eqnarray}

\noindent\textbf{Global Alignment.} Global alignment reflects the overall semantic similarity of modalities. We define a global similarity score with equation:
\begin{equation}
    \begin{split}
   S_G(R,T)=\frac{(c_{Ru}\cdot c_{Tu})}{||c_{Ru}||\cdot||c_{Tu}||},\label{sg}
    \end{split}
\end{equation}
where $R$ and $T$ represens an image and a text.

\noindent\textbf{Local Alignment.} Local alignment reflects the local similarity of modalities. 
We use Eq.~\eqref{A} to calculate the similarity matrix of the region-word and get cross-modal representation according to Eq.~\eqref{f*} and Eq.~\eqref{p*}.
\begin{eqnarray}
    A=(W_rF_{u})(W_tP_{u})^T,~~A\in \mathbb{R}^{n\times m} \label{A};\\
    f^*_i=\sum_{j=1}^m\alpha_{ij}p_j,~~\alpha_{ij}=\frac{exp(A_{ij})}{\sum_{j=1}^m exp(A_{ij})} \label{f*};\\
    p^*_j=\sum^n_{i=1}\beta_{ij}f_i,~~\beta_{ij}=\frac{exp(A_{ij})}{\sum_{i=1}^nexp(A_{ij})} \label{p*}.
\end{eqnarray}

Then, we define a local similarity score as 
\begin{equation}
    \begin{split}
    S_L(R,T)=\frac{1}{n}\sum^n_{i=1}\frac{f_{ui}\cdot f^*_i}{||f_{ui}||\cdot||f^*_i||}+\frac{1}{m}\sum^m_{j=1}\frac{p_{uj}\cdot p^*_j}{||p_{uj}||\cdot||p^*_j||}.
    \end{split}
\end{equation}

\subsection{The Overall Loss Function}
We define similarity score $S(R,T)=S_G(R,T)+\delta \cdot S_L(R,T)$, where $\delta$ is a hyperparameter that balances the importance of local similarity and global similarity. The loss function adopts a bidirectional triplet ranking loss as 
\begin{equation}
\begin{split}
\mathcal{L}=max\left[0,m-S(R,T)+S(R,\hat{T})\right]\\+max\left[0,m-S(R,T)+S(\hat{R},T)\right],
\end{split}
\end{equation}
 where $m$ is a pre-set margin, $(R,T)$ is the matched image-text pair, and $(\hat{R},\hat{T})$ are the hard negatives. 
 
 In a minibatch, the negative examples with the largest difference (instead of all negative examples) are used for training, i.e., $\hat{R}=argmax_{x!=R}S(x,T)$ and $\hat{T}=argmax_{y!=T}S(R,y)$.

\section{Experiments}

\subsection{Datasets and Settings}
To testify the effectiveness of our proposed method, we do experiments on Flickr30K~\cite{plummer2015flickr30k} and MSCOCO~\cite{lin2014microsoft}. Specifically,
Flickr30K has 31,000 images and 155,000 sentences, divided into 29,000 training images, 1,000 validation images and 1,000 test images. 
MSCOCO contains 123,287 images and 616,435 sentences, divided into 113,287 images for training, 5,000 for validation, and 5,000 for testing.

The commonly used metric for image-text retrieval is recall@K (abbreviated as R@K, K=1, 5, 10). Besides, to show the overall performance, we further calculate the $rSum$ as 
\begin{equation}
    rSum=\underbrace{R@1+R@5+R@10}_{Image~as~query}+\underbrace{R@1+R@5+R@10}_{Text~as~query}.
\end{equation}

We compared SGFN with quite a few SOTA methods including: 
MMCA~\cite{wei2020multi}, 
SHAN~\cite{ijcai2021p106}, CAAN~\cite{zhang2020context}, PFAN~\cite{ijcai2019p526}, 
GSMN~\cite{liu2020graph}, SGRAF~\cite{diao2021similarity}, SGM~\cite{wang2020cross}, 
MFA~\cite{li2021multi} and SCAN~\cite{lee2018stacked}.

 Our SGFN is trained with PyTorch on Nvidia TITAN XP. For the visual modality, the feature vector of the image region is extracted by Faster-RCNN to obtain 36 salient regions. Through a fully connected layer, the dimension is converted from 2048 to 1024. For the text modality, the dimension of the word vector is set to 300, and the dimension of word embedding 1024 is obtained through Bi-GRU. In the hierarchical attention fusion, $K$ takes 8, $\alpha$ takes 5 and $\beta$ takes 0. In the cross-modal fusion, we use a pre-trained Transformer encoder with 12 attention layers as the framework, with 16 heads and 768 hidden units. The epochs are 30 on Flickr30K and 20 on MSCOCO. The minibatch size of the Adam optimiser is 64. The initial learning rate for the first 15 epochs on Flickr30K is 0.0002, and the decay coefficient for subsequent epochs is 0.1. 
 The initial learning rate for the first 10 epochs on MSCOCO is 0.0005, and the decay coefficient for successive epochs is 0.1. 
The margin of hinge triplet loss is set to 0.2.
 
\subsection{Experimental Results}

\begin{table*}[!ht]
    \setlength\tabcolsep{4pt}
    \renewcommand{\arraystretch}{0.85}
 \begin{center}
 \caption{Performance comparisons of different methods w.r.t. R@K on Flickr30K and MSCOCO.} \label{table1}
    \begin{tabular}{c|c|c|c|c|c|c|c|c|c|c|c|c|c|c}
    \hline
   
    \multirow{3}{*}{Methods} & \multicolumn{7}{c|}{Flickr30K} &\multicolumn{7}{c}{MSCOCO} \\
    \cline{2-15}
    &\multicolumn{3}{c|}{Image-to-Text} & \multicolumn{3}{c|}{Text-to-Image} & \multirow{2}{*}{rSum} & \multicolumn{3}{c|}{Image-to-Text} & \multicolumn{3}{c|}{Text-to-Image} & \multirow{2}{*}{rSum}\\
    \cline{2-7}\cline{9-14}
    & R@1 &R@5 & R@10 & R@1 & R@5 & R@10& &R@1 & R@5 & R@10 & R@1 & R@5 & R@10 & \\
    \hline
    GSMN (2020) & 76.4 & 94.3 & 97.3 & 57.3 & 82.3 & 89.0 & 496.8 & 78.4 & 96.4 & 98.6 & 63.3 & 90.1 & 95.7 & 522.5\\
    MMCA (2020) & 74.2 & 92.8 & 96.4 & 54.8 & 81.4 & 87.8 & 487.4 & 74.8 & 95.6 & 97.7 & 61.6 & 89.8 & 95.2 & 514.7\\
    PFAN (2019b) & 70.0 & 91.8 & 95.0 & 50.4 & 78.7 & 86.1 & 472.0 & 76.5 & 96.3 & 99.0 & 61.6 & 89.6 & 95.2 & 518.2\\
    SACN (2018) & 67.4 & 90.3 & 95.8 & 48.6 & 77.7 & 85.2 & 465.0 & 72.7 & 94.8 & 98.4 & 58.8 & 88.4 & 94.8 & 507.9\\
    SGM (2020) & 71.8 & 91.7 & 95.5 & 53.5 & 79.6 & 86.5 & 478.6  & 73.4 & 93.8 & 97.8 & 57.5 & 87.3 & 94.3 & 504.1\\
    SHAN (2021) & 74.6 & 93.5 & 96.9 & 55.3 & 81.3 & 88.4 & 490.0 & 76.8 & 96.3 & 98.7 & 62.6 & 89.6 & 95.8 & 519.8\\
    CAAN (2020) & 70.1 & 91.6 & 97.2 & 52.8 & 79.0 & 87.9 & 478.6 & 75.5 & 95.4 & 98.5 & 61.3 & 89.7 & 95.2 & 515.6\\
    MFA (2021) & 77.2 & 94.6 & 97.4 & 57.4 & 82.3 & 89.3 & 498.2 & 76.8 & 95.8 & 98.8 & 61.8 & 89.7 & 95.5 & 518.0\\
    SGRAF (2021) & 77.8 & 94.1 & 97.4 & \textbf{58.5} & \textbf{83.0} & 88.8 & 499.6 & 79.6 & 96.2 & 98.5 & \textbf{63.4} & \textbf{90.7} & 96.1 & 524.5\\
    
    \hline
    SGFN & \textbf{81.5} & \textbf{97.6} & \textbf{98.4} & 58.2 & 82.5 & \textbf{91.8} & \textbf{510.0} & \textbf{81.1} & \textbf{96.9} & \textbf{99.0} & 63.1 & 90.2 & \textbf{97.1} & \textbf{527.4}\\
    \hline
    \end{tabular}
 \end{center}
\end{table*}

The results on Flickr30K are shown in Table~\ref{table1}. Our method ranks the top w.r.t. R@1, R@5 and R@10 on I2T (Image-to-Text) retrieval. From the perspective of mining modal extra information: PFAN and GSMN model the relative position information between regions through the region's physical position on the image. In contrast, our approach uses the scene graph to integrate extra information (attributes, relationships, etc.) into feature vectors. It can be seen that our approach outperforms PFAN and GSMN by 11.5\% and 5.1\% in terms of R@1. 
Besides, from the perspective of exploring the way of cross-modal fusion: When MMCA and SHAN perform cross-modal fusion, the global information and local information are fused separately. But our model combines the two. Our approach outperforms them by 7.3\% and 6.8\% in terms of R@1 on I2T retrieval. For T2I (Text-to-Image) retrieval, Our SGFN is close to the SOTA model w.r.t. R@1 and R@5 metrics and has a 2.5\% improvement compared to MFA on R@10. In addition, our model is 10.4\% higher than SGRAF in terms of rSum. Overall, the above results show that integrating modal semantic information into feature vectors and using global information to guide the comparison of local information can effectively improve the recall rate.

The results on a larger dataset MSCOCO are displayed in Table~\ref{table1}. We can observe that the performance of our method on I2T and T2I retrieval is roughly the same as the trend presented on Flickr30K. However, it is worth noting that on I2T retrieval, the performance improvement effect is not noticeable compared with Flickr30K. The reason is probably that sentences are more complex and contain fewer attributes in MSCOCO, and the introduce of the scene graph cannot well model the semantic information in the textual modality.

\subsection{Ablation Studies}
\begin{table}[!ht]
    \setlength\tabcolsep{4pt}
    \renewcommand{\arraystretch}{0.85}
 \begin{center}
    \caption{Performance comparisons on Flickr30K.} \label{AL}
    \begin{tabular}{c|c|c|c|c}
    \hline
    \multirow{2}{*}{Method} & \multicolumn{2}{c|}{Image-to-Text} &\multicolumn{2}{c}{Text-to-Image} \\
    \cline{2-5}
    & R@1 & R@10 & R@1 & R@10 \\
    \hline
    SGFN-intraF & 53.5 & 80.3 & 41.6 & 78.5\\
    SGFN-crossF & 72.3 & 94.5 & 54.4 & 85.7\\
    \hline
    SGFN & \textbf{81.5} & \textbf{98.4} & \textbf{58.2} & \textbf{91.8}\\
    \hline
    \end{tabular}
 \end{center}

\end{table}

To explore the impact of SGFN's intra- and inter-modal fusions on image-text retrieval, we conduct ablation studies on Flickr30K. Concretely, we compare SGFN with two variants: 
(1) SGFN-intraF only implements intra-modal fusion; (2) SGFN-crossF only implements cross-modal fusion. As shown in Table~\ref{AL}, the performances of the full model (SGFN) are significantly superior to its two variants with only intra- or inter-modal semantic fusion. The results reveal that it is more important to eliminate the gap between different modal spaces so that the semantically similar images/texts are also similar in the common feature space.

\subsection{Curves of the Parameters $\alpha$ and $\beta$}
\begin{figure}[ht]
\centering
\includegraphics[width=1.0\columnwidth]{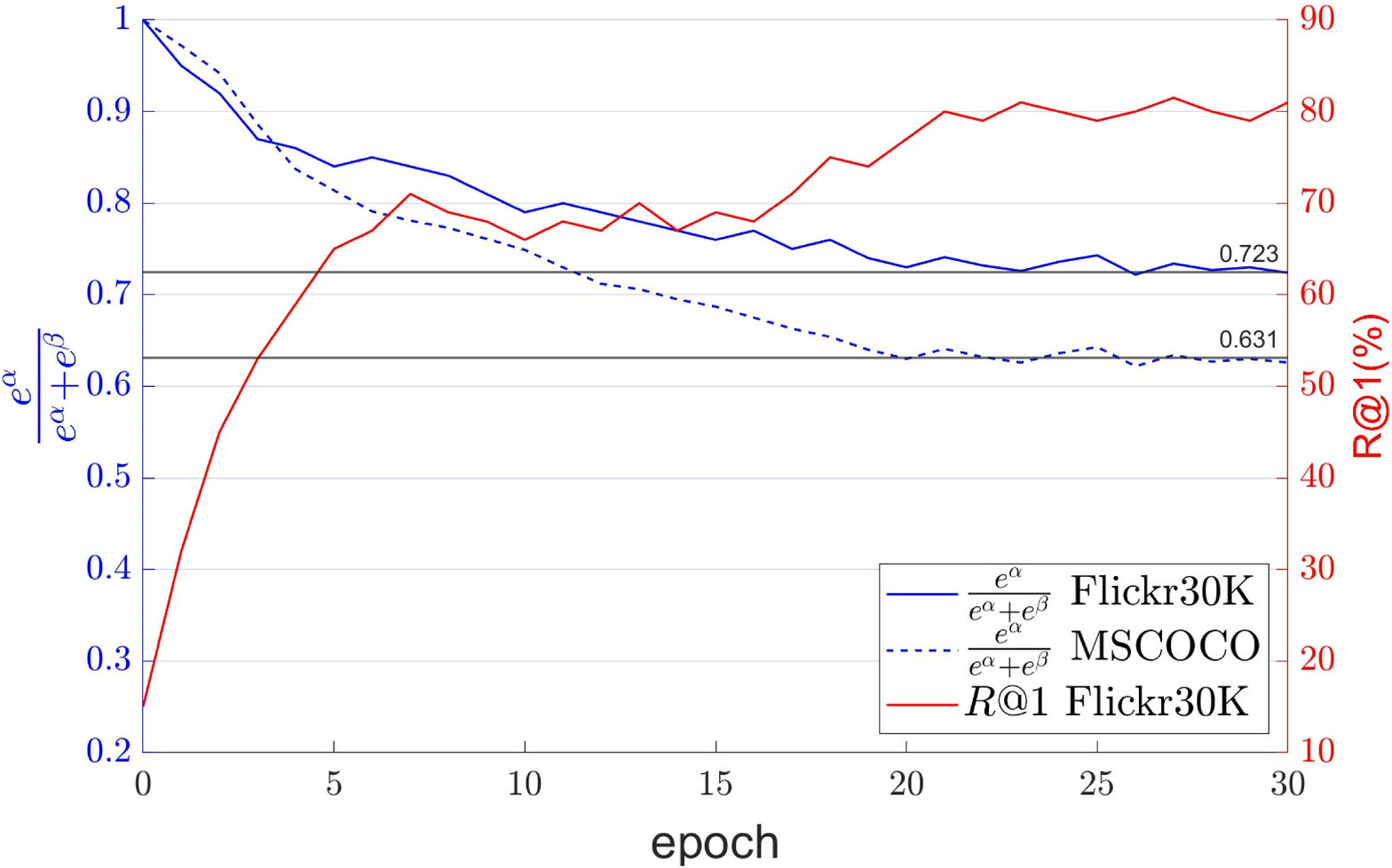}
\caption{The blue parts plot the variation of $\frac{e^{\alpha}}{e^{\alpha}+e^{\beta}}$ with epoch. The red part is the curve of R@1 with epoch.}
\label{figure3}
\end{figure}
To study the influence of the parameters $\alpha$ and $\beta$, we record their variations on I2T retrieval. As is shown in Fig.~\ref{figure3}, R@1 is better when $\frac{e^{\alpha}}{e^{\alpha}+e^{\beta}}$ is around 0.723 on Flickr30K, while R@1 is better when $\frac{e^{\alpha}}{e^{\alpha}+e^{\beta}}$ is around 0.631 on MSCOCO. 
A possible reason is that there is less semantic information (i.e., attributes or relations) in sentences on MSCOCO, so a larger proportion of  the attribute layer and relation layer is required than that on Flickr30K.

\subsection{Impact of the Hyperparameter $\delta$}
\begin{table}[!ht]
\renewcommand{\arraystretch}{0.9}
 \begin{center}
    \caption{Performance comparisons w.r.t. $\delta$ on Flickr30K.} \label{table3}
    \begin{tabular}{c|c|c|c|c}
    \hline
    \multirow{2}{*}{Model} & \multicolumn{2}{c|}{Image-to-Text} &\multicolumn{2}{c}{Text-to-Image} \\
    \cline{2-5}
    & R@1 & R@10 & R@1 & R@10 \\
    \hline
    $\delta$=0.0 & 70.4 & 85.7 & 48.6 & 81.3\\
    $\delta$=0.3 & \textbf{81.5} & \textbf{98.4} & \textbf{58.2} & \textbf{91.8}\\
    $\delta$=0.5 & 79.6 & 97.5 & 56.7 & 87.2\\
    $\delta$=0.7 & 77.4 & 95.3 & 55.3 & 85.6\\
    \hline
    \end{tabular}
 \end{center}
\end{table}

$\delta$ acts as a balancer, controlling the importance of global and local alignment on the final matching score. The performance comparisons w.r.t. different $\delta$ values are collected in Table~\ref{table3}. 
Clearly, We get the best performance when $\delta$ is set around 0.3. 
It's also worth noting that if $\delta$ is zero, SGFN only considers global alignment, resulting in performance degradation. Besides, a larger value of $\delta$ also has a negative impact on the results, which may cause the information obtained from global alignment to be obscured.

\subsection{A Visual Case on Region-Word Correspondence}

\begin{figure}[ht]
\centering
\includegraphics[width=1.0\columnwidth]{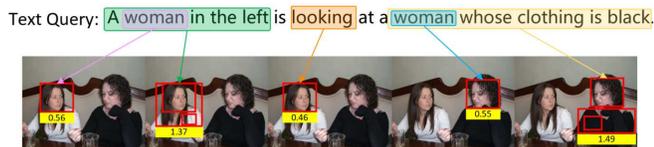}
\caption{Illustration of the correspondence between image regions and textual words after semantic enhancements, where the red box is the region with the highest similarity score.}
\label{figure4}
\end{figure}

We calculate the local similarities between the image/text feature matrices obtained after intra- and cross-modal fusions. We sort and visualize the region-word pairs according to their similarity scores, as shown in Fig.~\ref{figure4}. It can be seen that there is a significant distinction between the two regions ``\textit{woman}'' and the two words ``\textit{woman}''.

\section{Conclusion}

This paper mainly puts forward a novel image-text retrieval method (named SGFN) by using a hierarchical attention over image/text's scene graphs for intra-modal fusion and contextual vectors for inter-modal fusion in a unified neural network. In what follows, it defines the similarity score by global alignment and local alignment in a common space and minimises the hard negative based triplet loss to constrain the image/text representations.
Rich experients show that SGFN could yield better retrieval results than other SOTA methods, and ablation studies further testify the two fusion components do make a big difference for cross-modal retrieval.

\small
\bibliographystyle{IEEEbib}
\bibliography{reference}

\begin{thebibliography}{10}

\bibitem{chen2021crossvit}
Chun-Fu~Richard Chen, Quanfu Fan, and Rameswar Panda.
\newblock Crossvit: Cross-attention multi-scale vision transformer for image
  classification.
\newblock In {\em ICCV}, pages 357--366, 2021.

\bibitem{cong2022reltr}
Yuren Cong, Michael~Ying Yang, and Bodo Rosenhahn.
\newblock Reltr: Relation transformer for scene graph generation.
\newblock {\em arXiv preprint arXiv:2201.11460}, 2022.

\bibitem{diao2021similarity}
Haiwen Diao, Ying Zhang, Lin Ma, and Huchuan Lu.
\newblock Similarity reasoning and filtration for image-text matching.
\newblock In {\em AAAI}, volume~35, pages 1218--1226, 2021.

\bibitem{ijcai2021p106}
Zhong Ji, Kexin Chen, and Haoran Wang.
\newblock Step-wise hierarchical alignment network for image-text matching.
\newblock In {\em IJCAI}, pages 765--771, 2021.

\bibitem{lee2018stacked}
Kuang-Huei Lee, Xi~Chen, Gang Hua, Houdong Hu, and Xiaodong He.
\newblock Stacked cross attention for image-text matching.
\newblock In {\em ECCV}, pages 201--216, 2018.

\bibitem{Li2022VisionLanguageIT}
Feng Li, Hao Zhang, Yi-Fan Zhang, Shi~Tong Liu, Jian Guo, Lionel~M. Ni,
  Pengchuan Zhang, and Lei Zhang.
\newblock Vision-language intelligence: Tasks, representation learning, and
  large models.
\newblock {\em ArXiv}, abs/2203.01922, 2022.

\bibitem{li2021multi}
Wenhui Li, Yan Wang, Yuting Su, Xuanya Li, Anan Liu, and Yongdong Zhang.
\newblock Multi-scale fine-grained alignments for image and sentence matching.
\newblock {\em IEEE Trans Multimedia}, 2021.

\bibitem{lin2014microsoft}
Tsung-Yi Lin, Michael Maire, Serge Belongie, James Hays, Pietro Perona, Deva
  Ramanan, Piotr Doll{\'a}r, and C~Lawrence Zitnick.
\newblock Microsoft coco: Common objects in context.
\newblock In {\em ECCV}, pages 740--755, 2014.

\bibitem{liu2020graph}
Chunxiao Liu, Zhendong Mao, Tianzhu Zhang, Hongtao Xie, Bin Wang, and Yongdong
  Zhang.
\newblock Graph structured network for image-text matching.
\newblock In {\em CVPR}, pages 10921--10930, 2020.

\bibitem{nagrani2021attention}
Arsha Nagrani, Shan Yang, Anurag Arnab, Aren Jansen, Cordelia Schmid, and Chen
  Sun.
\newblock Attention bottlenecks for multimodal fusion.
\newblock {\em NeurIPS}, 34:14200--14213, 2021.

\bibitem{Nguyen2021ADL}
Manh-Duy Nguyen, Binh~T. Nguyen, and Cathal Gurrin.
\newblock A deep local and global scene-graph matching for image-text
  retrieval.
\newblock {\em ArXiv}, abs/2106.02400, 2021.

\bibitem{plummer2015flickr30k}
Bryan~A Plummer, Liwei Wang, Chris~M Cervantes, Juan~C Caicedo, Julia
  Hockenmaier, and Svetlana Lazebnik.
\newblock Flickr30k entities: Collecting region-to-phrase correspondences for
  richer image-to-sentence models.
\newblock In {\em ICCV}, pages 2641--2649, 2015.

\bibitem{radford2021learning}
Alec Radford, Jong~Wook Kim, Chris Hallacy, Aditya Ramesh, Gabriel Goh,
  Sandhini Agarwal, Girish Sastry, Amanda Askell, Pamela Mishkin, Jack Clark,
  et~al.
\newblock Learning transferable visual models from natural language
  supervision.
\newblock In {\em ICML}, pages 8748--8763. PMLR, 2021.

\bibitem{wang2018learning}
Liwei Wang, Yin Li, Jing Huang, and Svetlana Lazebnik.
\newblock Learning two-branch neural networks for image-text matching tasks.
\newblock {\em TPAMI}, 41(2):394--407, 2018.

\bibitem{wang2020cross}
Sijin Wang, Ruiping Wang, Ziwei Yao, Shiguang Shan, and Xilin Chen.
\newblock Cross-modal scene graph matching for relationship-aware image-text
  retrieval.
\newblock In {\em WACV}, pages 1508--1517, 2020.

\bibitem{ijcai2019p526}
Yaxiong Wang, Hao Yang, Xueming Qian, Lin Ma, Jing Lu, Biao Li, and Xin Fan.
\newblock Position focused attention network for image-text matching.
\newblock In {\em IJCAI}, pages 3792--3798, 2019.

\bibitem{wei2020multi}
Xi~Wei, Tianzhu Zhang, Yan Li, Yongdong Zhang, and Feng Wu.
\newblock Multi-modality cross attention network for image and sentence
  matching.
\newblock In {\em CVPR}, pages 10941--10950, 2020.

\bibitem{wu2019learning}
Yiling Wu, Shuhui Wang, Guoli Song, and Qingming Huang.
\newblock Learning fragment self-attention embeddings for image-text matching.
\newblock In {\em ACM MM}, pages 2088--2096, 2019.

\bibitem{zhang2020context}
Qi~Zhang, Zhen Lei, Zhaoxiang Zhang, and Stan~Z Li.
\newblock Context-aware attention network for image-text retrieval.
\newblock In {\em CVPR}, pages 3536--3545, 2020.

\bibitem{zhou2021deepvit}
Daquan Zhou, Bingyi Kang, Xiaojie Jin, Linjie Yang, Xiaochen Lian, Zihang
  Jiang, Qibin Hou, and Jiashi Feng.
\newblock Deepvit: Towards deeper vision transformer.
\newblock In {\em CVPR}, 2021.

\end{thebibliography}

\end{document}